%% file: main_arXiv.tex
\documentclass[11pt,letterpaper]{article}

\usepackage[textwidth=6in,textheight=8.5in,centering]{geometry}
\usepackage{amsmath,amssymb,amsthm}
\usepackage{graphicx}
\usepackage[round,authoryear]{natbib}
\usepackage{times}
\usepackage[T1]{fontenc}
\usepackage{placeins,booktabs}
\usepackage[normalem]{ulem}
\usepackage{microtype,comment,xcolor}
\usepackage{algorithm,algpseudocode}
\usepackage[hidelinks]{hyperref}

\newtheorem{theorem}{Theorem}
\newtheorem{proposition}[theorem]{Proposition}
\newtheorem{corollary}[theorem]{Corollary}
\theoremstyle{definition}
\newtheorem{definition}[theorem]{Definition}
\theoremstyle{remark}
\newtheorem{remark}[theorem]{Remark}

\newenvironment{keywords}{\par\smallskip\noindent\textbf{Keywords: }\ignorespaces}{\par}

\newcommand{\R}{\mathbb{R}}
\newcommand{\E}{\mathbb{E}}
\newcommand{\N}{\mathcal{N}}
\newcommand{\GP}{\mathcal{GP}}
\newcommand{\cH}{\mathcal{H}}
\newcommand{\cX}{\mathcal{X}}
\newcommand{\cD}{\mathcal{D}}
\newcommand{\Lop}{\mathcal{L}}

\DeclareMathOperator{\Cov}{Cov}

\newcommand{\dd}{\,\mathrm{d}}
\newcommand{\kin}{\mathrm{kin}}

\title{Learning Gaussian Process Priors with\\Linear Semigroups and Kinetic Regularisation}
\author{Angela Lavinia Varazi\thanks{\texttt{angela.lavinia.varazi@gmail.com}} \and
        Felipe Tobar\thanks{\texttt{f.tobar@imperial.ac.uk}}}
\date{\small Department of Mathematics, Imperial College London}
\begin{document}
\maketitle

\input{sections/abstract}

\input{sections/introduction}
\input{sections/background}
\input{sections/semigroup-priors}

\input{sections/learning}
\input{sections/applications}
\input{sections/conclusion}

\FloatBarrier
\newpage
\bibliography{references}
\clearpage
\input{sections/appendix}

\end{document}

%% file: sections/abstract.tex
\begin{abstract}
We introduce a semigroup-based construction of Gaussian process (GP) priors. A linear semigroup transforms a source GP at $t=0$ into a prior at $t=1$, enabling computationally tractable regression. Linearity preserves Gaussianity, allowing us to characterise the evolution of the covariance and draw connections with optimal transport and convolution models. The semigroup is parametrised by its generator, which is learned through two competing objectives: the marginal likelihood encourages agreement with observations, while a kinetic-energy regulariser favours lower-energy trajectories and limits unnecessary mass displacement from the source GP. Experiments on independent GP draws show that, with sufficiently many training functions, the learned prior provides accurate predictions on held-out functions, while kinetic regularisation substantially reduces the energy of the generated trajectories without compromising prediction accuracy.
\end{abstract}

\begin{keywords}%
  Gaussian processes, kernel learning, linear semigroups, kinetic regularisation%
\end{keywords}

%% file: sections/introduction.tex

\section{Introduction}

Gaussian processes (GPs) are Bayesian nonparametric priors over continuous-time functions, widely used in regression \citep{rasmussen2006}. The covariance of the GP determines the classes of functions promoted under the chosen prior, allowing the GP to model a wide variety of temporal features, including smoothness, periodicity, and  long-range correlation. However, the ability of the GP to incorporate expert knowledge also comes with the challenge of having to choose the covariance structure; this becomes particularly relevant when  little information about the data-generating system is available, or when such information cannot be practically encoded into the kernel.

Kernel design has long been a central direction in the GP literature. In the stationary case,  covariance design follows combinations of elementary kernels \citep{duvenaud2013}, spectral mixture models \citep{wilson2013}, nonparametric convolutions \citep{tobar2015}, and composing kernels with neural networks \citep{wilson2016}. In the nonstationary setting, standard paradigms for covariance design include process-convolution and locally varying covariance constructions \citep{higdon1998,paciorek2006}, nonstationary spectral kernels \citep{remes2017}, as well as GP change-point models \citep{saatci2010}.

Despite this rich literature, constructing flexible GP priors remains challenging, particularly when one seeks to avoid having to specify a highly structured covariance function involving several non-interpretable parameters. Inspired by recent developments in flow-based generative modelling, we propose a construction for expressive GP priors by evolving a source GP through an ordinary differential equation. The key distinction of our method with respect to current flow-based models is threefold: first, we consider a \emph{semigroup}, meaning that the linear velocity field is straightforward to interpret and parametrise; second, our approach is functional and thus induces a transport on the space of continuous-time functions; third, we introduce a kinetic-energy regulariser that favours lower-energy trajectories while maintaining agreement with the observations. 


Our construction treats GP prior design as learning a transformation of a source process, with explicit control over both the resulting covariance and the energy required to reach it.

%% file: sections/background.tex
\section{Background and related work}
\label{sec:background}

\subsection{Transformations of Gaussian processes}
\label{sec:background-gps}

A GP is a collection of random variables
$\{f(x):x\in\cX\}$ such that every finite collection is jointly Gaussian. Fully characterised by its mean function
$m(x)=\E[f(x)]$ and covariance function
$k(x,x')=\Cov(f(x),f(x'))$, a GP is denoted by
$f\sim\mathcal{GP}(m,k)$. We refer the reader to \cite{rasmussen2006} for a thorough presentation of likelihood-based GP training (on which our training strategy builds) and closed-form posterior predictive distributions.

Linear transformations of GPs provide a mechanism for constructing covariance functions. Consider a zero-mean GP $Z_0$ with covariance $k_0$ and
a deterministic integral operator
\[
(Tf)(x)=\int_{\cX}h(x,u)f(u)\,\dd u,
\]
where $h$ is a suitable filter. The transformed process $Z=TZ_0$ is again a zero-mean GP. Location-dependent filters allow nonstationary models, whereas translation-invariant filters preserve stationarity~\citep{paciorek2006}. Process-convolution models
\citep{higdon1998,tobar2015} use Gaussian white noise as the source, with the transformation interpreted as a stochastic integral. Our work is motivated by these constructions, but instead of a single-step convolutional transformation, we evolve the source GP through a linear semigroup.

\subsection{Semigroups, generators and kinetic energy}
\label{sec:background-semigroups}

\begin{definition}[Semigroup and generator]A family $(T_t)_{t\in[0,1]}\subset\Lop(\cH)$ is a strongly continuous
semigroup if, for $t,s\in[0,1]$ such that $t+s\leq1$,
\begin{equation}
T_0=I,\qquad T_{t+s}=T_tT_s,
\label{eq:c0}
\end{equation}
and, for any $f\in\cH$,
$\lim_{t\downarrow0}\|T_tf-f\|_\cH=0.$
The generator of the semigroup is defined by
\begin{equation}
Af=\lim_{h\downarrow0}\frac{T_hf-f}{h},
\qquad f\in \cD(A),
\label{eq:generator}
\end{equation}
where $\cD(A)\subseteq\cH$ is the set of elements for which this limit exists.
\end{definition}

The generator is densely defined on its domain and uniquely
determines the semigroup. For $f\in\cD(A)$,
the evolution satisfies~\citep{engel2000}
\begin{equation}
\frac{\dd}{\dd t}T_tf=AT_tf=T_tAf.
\label{eq:semigroup-evolution}
\end{equation}
When $A$ is bounded, the semigroup admits the
operator-exponential representation $T_t=e^{tA}$. For any bounded linear operator $T$ on $\cH$, its adjoint $T^*$
satisfies $\langle Tf,g\rangle_\cH=\langle f,T^*g\rangle_\cH$.

In the generative setup considered in this work, under suitable regularity and integrability conditions, the generator determines the velocity $\dot Z_t=AZ_t$. This motivates the use of the expected kinetic-energy functional
\begin{equation}
\mathcal L_{\mathrm{kin}}
=\frac12\int_0^1\E\|AZ_t\|_\cH^2\,\dd t,
\label{eq:kinetic-energy-intro}
\end{equation}
as a penalisation of rapid motion along the semigroup's evolution, thus providing a means to regularise the generating path across different alternative trajectories reaching the same endpoint law.
\subsection{Related work}
\label{sec:related-work}

Our approach relates to operator-based covariance constructions, including process convolution \citep{higdon1998,tobar2015}, diffusion kernels \citep{kondor2002}, and stochastic partial differential equation models \citep{lindgren2011}. We learn a semigroup generator that determines both the endpoint covariance and its regularised generating evolution. Our finite Fourier implementation also uses a truncated basis representation, as in \citet{solinsarkka2020}, with the coefficient covariance learned through the generator.

Kinetic-energy regularisation has also been explored
for continuous-time generative dynamics, including
neural ordinary differential equations
\citep{finlay2020}. In our setting, the penalty is
the expected squared function-space velocity under
a fixed Gaussian initial law. Unlike general nonlinear
flows, the linear evolution preserves Gaussianity,
so the endpoint likelihood and posterior predictions
remain analytically tractable through standard GP
inference. Furthermore, because different generators
may induce the same endpoint law, the kinetic-energy
penalty provides a criterion for selecting among
otherwise indistinguishable evolutions, without
assuming that the generating dynamics are identifiable.

%% file: sections/semigroup-priors.tex
\section{Semigroup priors}
\label{sec:priors}

\subsection{Setup}
\label{sec:prior-setup}

We consider a domain $\cX\subseteq\R^d$ and a real separable
Hilbert space $\cH$ of functions $f:\cX\to\R$ for which point
evaluation is continuous. In particular, for each $x\in\cX$ there
exists a Riesz representer $r_x\in\cH$ such that
$f(x)=\langle f,r_x\rangle_\cH$~\citep{aronszajn1950}.
Also, recall that a random element $Z\in\cH$ is Gaussian if
$\langle Z,h\rangle_\cH$ is Gaussian for every $h\in\cH$
\citep{daprato2002}; its mean $m\in\cH$ and covariance
operator $C:\cH\to\cH$ are defined by
\[
m=\E[Z],\qquad
Ch=\E\big[\langle Z-m,h\rangle_\cH(Z-m)\big].
\]
Thus, $\Cov(\langle Z,h\rangle_\cH,\langle Z,g\rangle_\cH)
=\langle Cg,h\rangle_\cH$.
Throughout, $t\in[0,1]$ denotes auxiliary
\emph{generative time}, distinct from the input $x\in\cX$, which may represent \emph{physical time} in some applications.

\subsection{Semigroups and Gaussian evolution}

\begin{definition}[Semigroup prior]\label{def:semigroup-priors}
Let $Z_0\sim\N(m_0,C_0)$ be a Gaussian random element of $\cH$,
where $m_0\in \cH$ and $C_0$ a positive semidefinite, self-adjoint, trace-class covariance operator. Consider a strongly continuous semigroup $(T_t)_{t\in[0,1]}$ with generator $A$. Define the trajectory $(Z_t)_{0\leq t\leq1}$ describing the evolution from the initial (source) process $Z_0$ to the terminal one $Z_1$, where
\[
Z_t=T_tZ_0,\qquad t\in[0,1].
\]
The \emph{semigroup prior} is the law of the endpoint process 
\(
Z_1=T_1Z_0.
\)

\end{definition}

The endpoint $Z_1$ defines the prior used for regression, while the
intermediate random functions describe how this prior is generated.

\begin{proposition}[Gaussianity and covariance evolution]\label{prop:gaussian}
The trajectory $(Z_t)_{t\in[0,1]}$ is jointly Gaussian,
in the sense that for any finite collection of times
$t_1,\ldots,t_n\in[0,1]$ and elements
$h_1,\ldots,h_n\in\cH$, the random vector
\[
\left(
\langle Z_{t_1},h_1\rangle_\cH,\ldots,
\langle Z_{t_n},h_n\rangle_\cH
\right)
\]
is multivariate Gaussian. In particular, $Z_t\sim\N(m_t,C_t)$, where
\begin{equation}
m_t=T_tm_0,\qquad
C_{t,s}=T_tC_0T_s^*,
\label{eq:moments}
\end{equation}
with $C_{t,s}=\Cov(Z_t,Z_s)$ and $C_t=C_{t,t}$ being positive semidefinite,
self-adjoint and trace class.
\end{proposition}

\begin{proof}
For any finite collection of times $t_i$ and elements
$h_i\in\cH$, we have
\(
\sum_i \langle Z_{t_i},h_i\rangle_\cH
=
\left\langle Z_0,\sum_i T_{t_i}^*h_i
\right\rangle_\cH,
\)
which is Gaussian since $Z_0$ is a Gaussian random
element of $\cH$. Hence, $(Z_t)_{t\in[0,1]}$ is
jointly Gaussian.

By bounded linearity, $m_t=T_tm_0$. Furthermore,
for any $h,g\in\cH$,
\[
\Cov\left(
\langle Z_t,h\rangle_\cH,
\langle Z_s,g\rangle_\cH
\right) =
\left\langle C_0T_s^*g,T_t^*h
\right\rangle_\cH=
\left\langle T_tC_0T_s^*g,h
\right\rangle_\cH.
\]
Thus, $C_{t,s}=T_tC_0T_s^*$.
Since $C_0$ is positive semidefinite, self-adjoint
and trace class, so is $C_t=T_tC_0T_t^*$.
\end{proof}

\begin{corollary}[Covariance evolution]
\label{cor:covariance}
Under the assumptions of Proposition~\ref{prop:gaussian},
the covariance operator $C_t$ satisfies
\begin{equation}
C_{t+s}=T_tC_sT_t^*,
\qquad t,s\geq0,\quad t+s\leq1.
\label{eq:cov_semigroup}
\end{equation}
If $A$ is bounded, its evolution is governed by
\begin{equation}
\frac{\mathrm d}{\mathrm dt}C_t
=AC_t+C_tA^*,
\label{eq:cov_ode}
\end{equation}
where differentiation is taken in operator norm.
\end{corollary}

\begin{proof}
By the semigroup property and Proposition~\ref{prop:gaussian}, we have
\(
C_{t+s}
=T_tT_sC_0T_s^*T_t^*
=T_tC_sT_t^*.
\)
When $A$ is bounded, differentiating
$C_t=T_tC_0T_t^*$ yields the required expression for $\frac{\mathrm d}{\mathrm dt}C_t$.
\end{proof}

By the Riesz representation in Section~\ref{sec:prior-setup}, $Z_t$ defines a GP over $\mathcal X$ with mean $m_t(x)$ and covariance kernel
\begin{equation}
k_t(x,x')
=
\langle r_x,C_tr_{x'}\rangle_\cH
=
\langle r_x,T_tC_0T_t^*r_{x'}\rangle_\cH.
\label{eq:evaluation}
\end{equation}
The same construction determines the cross-covariances
\(
\Cov\bigl(Z_t(x),Z_s(x')\bigr)
=\langle r_x,C_{t,s}r_{x'}\rangle_\cH.
\)
Lastly, the terminal process satisfies
\[
Z_1\sim\mathcal{GP}(m_1,k_1),
\]
and can therefore be used as a prior for Gaussian
process regression.

\begin{remark}[White-noise initialisation]
A Gaussian white-noise source can also be used, although it is not
$\cH$-valued in infinite dimensions: the semigroup must make
$C_t=\sigma_\xi^2T_tT_t^*$ trace class for $t>0$~\citep{riedle2014}.
Kinetic regularisation then requires suitable regularity and integration
away from zero, over $[\varepsilon,1]$ with $\varepsilon>0$.
\end{remark}

\subsection{Finite-dimensional case}
\label{sec:special_cases}

For $\cH=\R^d$, the generator $A\in\R^{d\times d}$
defines the semigroup $T_t=e^{tA}$, inducing a
Gaussian probability path $p_t=\N(m_t,C_t)$ with
\[
m_t=e^{tA}m_0,\qquad
C_t=e^{tA}C_0e^{tA^\top}.
\]

Importantly, this construction can connect any two
positive-definite covariance matrices
$\bar C_0,\bar C_1\in\R^{d\times d}$.
Indeed, consider the symmetric positive definite matrix
\[
B=\bar C_0^{-1/2}
\left(\bar C_0^{1/2}\bar C_1\bar C_0^{1/2}\right)^{1/2}
\bar C_0^{-1/2},
\]
which satisfies
$B\bar C_0B^\top=\bar C_1$.
Choosing the generator $A=\log B$ yields
\[
C_t=B^t\bar C_0B^t,
\]
so that $C_0=\bar C_0$ and $C_1=\bar C_1$.
Thus, any two nondegenerate centred multivariate
Gaussian distributions can be connected by a
semigroup-induced Gaussian probability path.

\begin{remark}[Connection to optimal transport]
The matrix $B$ coincides with the optimal transport
map between the two centred Gaussian distributions
under quadratic cost. However, the semigroup
interpolation $B^t$ generally differs from the
Wasserstein geodesic interpolation $(1-t)I+tB$.
\end{remark}

%% file: sections/learning.tex
\section{Learning the semigroup}
\label{sec:model}

We parametrise a semigroup $T_t^\theta$ via its generator
$A_\theta$, where the parameter $\theta$ is learnt based on a loss defined below.  For a fixed source law $Z_0\sim\N(m_0,C_0)$, write
$Z_t^\theta=T_t^\theta Z_0$ and consider the following notation for the endpoint mean and kernel, respectively:
\begin{equation}
\begin{aligned}
m_\theta(x)&=(T_1^\theta m_0)(x),\\
k_\theta(x,x')&=\langle r_x,T_1^\theta C_0(T_1^\theta)^*r_{x'}\rangle_\cH.
\end{aligned}
\label{eq:model-kernel}
\end{equation}

\subsection{Observation model and learning objective}
\label{sec:model-objective}

Given and observation dataset $\cD=\{(x_i,y_i)\}_{i=1}^N$, assume
$y_i=Z_1^\theta(x_i)+\varepsilon_i$, with i.i.d.~
$\varepsilon_i\sim\N(0,\sigma_n^2)$, independent of $Z_0$.
Then, for $\Theta=(\theta,\sigma_n)$,
\begin{equation}
\mathbf y\mid\Theta\sim\N(\mathbf m_\theta,K_\Theta),\qquad
(K_\Theta)_{ij}=k_\theta(x_i,x_j)+\sigma_n^2\delta_{ij},
\label{eq:model-marginal}
\end{equation}
where $(\mathbf m_\theta)_i=m_\theta(x_i)$, and the negative log marginal likelihood (NLL) is
\begin{equation}
  \mathcal L_{\rm NLL}(\Theta)=\tfrac12(\mathbf y-\mathbf m_\theta)^\top K_\Theta^{-1}(\mathbf y-\mathbf m_\theta)
  +\tfrac12\log\det K_\Theta+\tfrac N2\log 2\pi.
  \label{eq:nll}
\end{equation}
We learn the semigroup by minimising the NLL regularised by the kinetic-energy penalty in~\eqref{eq:kinetic-energy-intro}:

\begin{equation}
\min_{\substack{\theta\in\Theta_A\\\sigma_n\geq\sigma_{\min}}}
\left\{\mathcal L_{\mathrm{NLL}}(\Theta)
+\lambda_{\kin}\mathcal L_{\kin}(\theta)\right\},
\qquad
\mathcal L_{\kin}(\theta)
=\frac12\E\int_0^1\|A_\theta T_t^\theta Z_0\|_\cH^2\,\dd t.
\label{eq:model-objective}
\end{equation}
Here $\lambda_{\kin}\geq0$ is the kinetic-energy regularisation weight,
$\sigma_{\min}>0$, and $\Theta_A$ restricts the generators to
mean-square absolutely continuous trajectories satisfying
$\dot Z_t^\theta=A_\theta Z_t^\theta$ almost everywhere, and finite
kinetic energy.

\begin{remark}[Role of kinetic regularisation]
Different generators can induce the same endpoint law through distinct
trajectories; the kinetic penalty favours lower-energy evolutions and
limits the displacement of the terminal prior from the initial one.
Indeed,
\[
W_{2,\cH}^2(\mu_0,\mu_1^\theta)
\leq
\mathbb E\|Z_1^\theta-Z_0\|_\cH^2
\leq 2\mathcal L_{\mathrm{kin}}(\theta),
\]
where $\mu_0$ and $\mu_1^\theta$ denote the initial
and terminal laws, respectively. The first inequality uses the coupling
$(Z_0,Z_1^\theta)$, and the second follows from Cauchy--Schwarz in time.
Thus, the penalty controls both the
energy of the generating trajectory and the squared $2$-Wasserstein
distance between the initial and terminal laws.
\label{rem:kinetic-role}
\end{remark}

\subsection{Choosing the regularisation weight}
\label{sec:model-weight}

\begin{remark}[Normalisation of the regularisation weight]
\label{rem:lambda-normalization}
Writing $\lambda_{\kin}=N\bar\lambda_{\kin}$ and dividing the objective
by $N$ gives the equivalent optimisation problem
\[
\min_{\Theta}\left\{
\frac1N\mathcal L_{\mathrm{NLL}}(\Theta)
+\bar\lambda_{\kin}\mathcal L_{\kin}(\theta)\right\},
\]
subject to the constraints in~\eqref{eq:model-objective}.
Thus, $\bar\lambda_{\kin}$ sets the penalty relative to NLL per observation.
This convention does not imply that information about the generator grows
linearly with $N$, since observations within a function are correlated.
\end{remark}

We search over the normalised weight $\bar\lambda_{\kin}$, tracing a trade-off curve between kinetic energy and NLL. We seek a knee where energy is reduced with only a small increase in NLL, beyond which further reductions incur a much larger likelihood cost. The experimental grid and selection rule are specified in Appendix~\ref{sec:weight-selection-details}.

\subsection{Loss computation and optimisation}
\label{sec:model-computation}

The Gaussian likelihood is evaluated directly from
$\mathbf m_\theta$ and $K_\Theta$ using Cholesky factorisation;
it integrates out the source and requires no sampled initial functions.
The kinetic expectation can likewise be computed in closed form when
the representation permits, leaving only the time integral.
Otherwise, we estimate the expectation using source draws held fixed during optimisation and apply numerical quadrature in time. The estimator and fitting procedure (Algorithm~\ref{alg:semigroup-learning}) are given in Appendix~\ref{sec:general-computation}.

%% file: sections/applications.tex

\section{Experimental validation}
\label{sec:applications}

\subsection{Model specification}
\label{sec:application-model}

\paragraph{Function space.} We instantiate the semigroup prior using a finite Fourier representation~\citep[cf.][]{solinsarkka2020} on  $\cX=[0,1)$. Consider the function space
$\cH_{13}=\operatorname{span}\{\phi_0,\ldots,\phi_{12}\}$,
equipped with the $L^2[0,1]$ inner product, where $\phi_0(x) = 1,$ and 
\[
\begin{aligned}
\phi_{2k-1}(x) &= \sqrt{2}\cos(2\pi kx),\\
\phi_{2k}(x) &= \sqrt{2}\sin(2\pi kx),
\qquad k=1,\ldots,6.
\end{aligned}
\]
Denoting $\phi(x)=(\phi_0(x),\ldots,\phi_{12}(x))^\top$, orthonormality implies
$\|\phi^\top c\|_{\cH_{13}}^2=\|c\|_2^2$, meaning that function-space norms can be evaluated directly from the coefficients $c$.

\paragraph{Source process.}
We define the initial random function as
$Z_0(x)=\phi(x)^\top c_0$, where
\[
c_0\sim\N(0,Q),\qquad
Q=\operatorname{diag}(q_0,\ldots,q_{12}).
\]
We choose the variances to favour low-frequency components and
normalise them so that
$\E\|Z_0\|_{\cH_{13}}^2=\operatorname{tr}(Q)=1$.
Equal variances within each sine--cosine pair yield
a stationary initial GP with
$\operatorname{Var}[Z_0(x)]=1$ for all $x\in\cX$.
To satisfy these requirements, we set
\[
q_j=\frac{w_j}{\sum_{\ell=0}^{12}w_\ell},
\qquad
w_j=\left[1+(\nu_j/2.3)^4\right]^{-1},
\qquad
(\nu_j)_{j=0}^{12}=(0,1,1,\ldots,6,6).
\]

\paragraph{Generator parametrisation.}
We represent the generator $A_\theta$ through a
matrix $B_\theta\in\R^{13\times13}$ acting on the
Fourier coefficients:
\[
A_\theta(\phi^\top c)=\phi^\top B_\theta c.
\]
We parametrise this matrix via the decomposition
\begin{equation}
B_\theta=S_\theta-R_\theta R_\theta^\top-\gamma I,
\qquad S_\theta^\top=-S_\theta,
\label{eq:fourier-generator}
\end{equation}
motivated by the following mechanisms: i) The skew-symmetric matrix $S_\theta$ rotates the Fourier coefficients maintaining their Euclidean norm; ii) The negative semidefinite component $-R_\theta R_\theta^\top$ introduces direction-dependent damping in coefficient space; and iii) the term $-\gamma I$ ensures uniform strict dissipation, we choose $\gamma=0.01$.

This parametrisation implies that, for any $f=\phi^\top c\in\cH_{13}$,
orthonormality gives
\[
\langle A_\theta f,f\rangle_{\cH_{13}}
=c^\top B_\theta c
=-\|R_\theta^\top c\|_2^2-\gamma\|c\|_2^2
\leq-\gamma\|f\|_{\cH_{13}}^2.
\]
Therefore, the generated semigroup satisfies
\(
\|T_t^\theta f\|_{\cH_{13}}
\leq e^{-\gamma t}\|f\|_{\cH_{13}},
\) meaning that the evolution contracts function-space norms.

\paragraph{Semigroup evolution and kinetic energy.}
In this finite-dimensional space, the semigroup is represented by
$e^{tB_\theta}$, giving the random function and coefficient covariance
\begin{equation}
Z_t^\theta(x)=\phi(x)^\top e^{tB_\theta}c_0,
\qquad
\Sigma_t^\theta=e^{tB_\theta}Qe^{tB_\theta^\top}.
\label{eq:fourier-evolution}
\end{equation}
The terminal process is a zero-mean GP with kernel $k_\theta$, and the
expected kinetic energy is obtained analytically under the initial Gaussian law:
\[
k_\theta(x,x')=\phi(x)^\top\Sigma_1^\theta\phi(x'),
\qquad
\mathcal L_{\mathrm{kin}}(\theta)
=\frac12\int_0^1
\left\|B_\theta e^{tB_\theta}Q^{1/2}\right\|_F^2\,\mathrm dt.
\]
The time integral is also available in closed form, $\mathcal{L}_{\mathrm{kin}}(\theta)=\tfrac{1}{2}\operatorname{tr}(B_\theta V_\theta B_\theta^\top)$, where $V_\theta=\int_0^1\Sigma^\theta_t\,\mathrm{d}t$ solves the Lyapunov equation $B_\theta V_\theta+V_\theta B_\theta^\top=\Sigma^\theta_1-Q$. Parameter constraints and numerical optimisation settings are detailed in Appendix~\ref{sec:implementation-details}.

\subsection{Prior recovery: experimental setup}
\label{sec:proposed-experimental-setup}

We test recovery of a shared GP prior from $M\in\{1,4,16,64,256\}$ independent function draws and assess the effect of kinetic regularisation. The target is a nonstationary transformation of the Fourier source in Section~\ref{sec:application-model}, using localised damping and mode coupling. Recovery concerns the endpoint law, and not the identification of its true generator. Each training function has 40 noisy observations outside the common gap $[0.34,0.49]$.

We learn a shared generator and noise level jointly across functions, using objective~\eqref{eq:model-objective} with the summed NLL normalised by $N=40M$. We compare knee-selected regularisation, unregularised learning, and the fixed source prior, for which only observation noise is fitted on the same data. The target specification, fitting procedure, and evaluation details are given in Appendix~\ref{sec:synthetic-specification}, with numerical optimisation settings in Appendix~\ref{sec:implementation-details}.

We evaluate prediction on independent held-out functions, conditioning on each function's own observations while keeping the learned generator and noise level fixed. Latent RMSE and noisy NLPD assess overall prediction, while gap RMSE assesses prediction within the unobserved interval. Results are aggregated over 20 random seeds, as detailed in Table~\ref{tab:proposed-multi-function-results}.

\subsection{Path selection and regularisation strength}
\label{sec:proposed-path-selection}

\begin{figure}[!t]
\centering
\includegraphics[width=\textwidth]{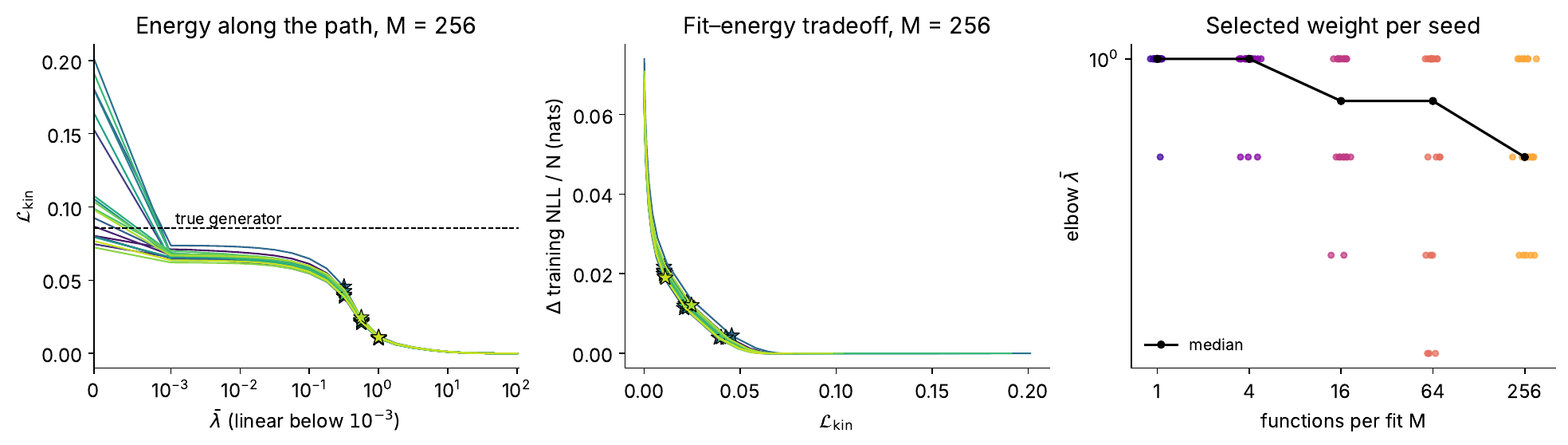}
\caption{Regularisation paths at $M=256$ (left and centre; colours denote seeds). Left: energy versus weight $\bar\lambda_{\kin}$; dashed line: true-generator energy. Centre: NLL increase per observation over the unregularised fit versus energy; stars: selected knees (referred to as \emph{elbow} in the plot). Right: selected weights $\bar\lambda_{\kin}$ and their median versus $M$. The left weight axis is linear below $10^{-3}$.}
\label{fig:proposed-multi-paths}
\end{figure}

Figure~\ref{fig:proposed-multi-paths} separates two effects of kinetic regularisation: selecting a lower-energy path with little change in likelihood, and trading likelihood fit for further energy reduction. For $M=256$, unregularised fits have energies ranging from $0.07$ to $0.20$ across seeds. A small weight, $\bar\lambda_{\kin}=10^{-3}$, narrows this range to $0.062$--$0.074$, while changing training NLL by at most $0.05$ nats in total. Thus, fits with nearly identical likelihoods can have substantially different path energies.

For $\bar\lambda_{\kin}$ between $10^{-3}$ and $10^{-1}$, energy remains near a plateau of $0.055$--$0.074$, below the true generator's energy of $0.086$. Beyond $\bar\lambda_{\kin}\approx0.3$, further energy reductions come with a more pronounced likelihood cost. The selected knees lie in this region ($\bar\lambda_{\kin}=0.32$--$1$), with energies of $0.010$--$0.046$ and NLL increases of $0.004$--$0.022$ nats per observation.

Although these per-observation increases are small, they amount to $40$--$222$ nats over the $10{,}240$ observations. Selecting a knee after rescaling both axes to $[0,1]$ does not enforce a small total likelihood loss. At $M=256$, the selected fits over-smooth the learned prior (Figure~\ref{fig:proposed-multi-prior}), illustrating this limitation of the selection rule. The selected $\bar\lambda_{\kin}$ nevertheless changes little with $M$ (Figure~\ref{fig:proposed-multi-paths}, right).

\subsection{Learning a prior from multiple endpoint functions}
\label{sec:proposed-multi-function}

\begin{table}[!t]
\centering
\caption{Held-out prediction versus training-set size $M$. Metrics average 50 functions per seed, then 20 seeds; $\pm$ denotes SD across seed averages. Energy is a mean and $\bar\lambda_{\kin}$ a median across seeds. The true model is an oracle reference. Evaluation details: Appendix~\ref{sec:synthetic-specification}.}
\label{tab:proposed-multi-function-results}
\footnotesize
\setlength{\tabcolsep}{4pt}
\begin{tabular}{@{}clccccc@{}}
\toprule
$M$ & Model & $\bar\lambda_{\kin}$ & $\mathcal L_{\kin}$ & Latent RMSE & Gap RMSE & Noisy NLPD \\
\midrule
1 & Knee-selected & 1 & 0.1369 & $0.2877\pm0.0411$ & 0.5470 & $2.9889$ \\
 & Unregularised & 0 & 0.6571 & $0.3743\pm0.0511$ & 0.7067 & $4.2575$ \\
 & Fixed source ($B=0$) & -- & 0.0000 & $0.1093\pm0.0076$ & 0.2366 & $-0.7471$ \\
\midrule
4 & Knee-selected & 1 & 0.1249 & $0.2848\pm0.0567$ & 0.5470 & $2.5624$ \\
 & Unregularised & 0 & 0.5156 & $0.3769\pm0.0352$ & 0.6236 & $7.5006$ \\
 & Fixed source ($B=0$) & -- & 0.0000 & $0.1078\pm0.0075$ & 0.2325 & $-0.7644$ \\
\midrule
16 & Knee-selected & 0.781 & 0.0440 & $0.1514\pm0.0242$ & 0.3298 & $-0.4382$ \\
 & Unregularised & 0 & 0.2493 & $0.1910\pm0.0370$ & 0.4269 & $0.3516$ \\
 & Fixed source ($B=0$) & -- & 0.0000 & $0.1073\pm0.0072$ & 0.2313 & $-0.7668$ \\
\midrule
64 & Knee-selected & 0.781 & 0.0286 & $0.1045\pm0.0083$ & 0.2240 & $-0.7647$ \\
 & Unregularised & 0 & 0.1881 & $0.1131\pm0.0097$ & 0.2476 & $-0.7032$ \\
 & Fixed source ($B=0$) & -- & 0.0000 & $0.1072\pm0.0072$ & 0.2310 & $-0.7677$ \\
\midrule
256 & Knee-selected & 0.562 & 0.0225 & $0.0965\pm0.0059$ & 0.2021 & $-0.7937$ \\
 & Unregularised & 0 & 0.1153 & $0.0971\pm0.0055$ & 0.2036 & $-0.7904$ \\
 & Fixed source ($B=0$) & -- & 0.0000 & $0.1073\pm0.0071$ & 0.2312 & $-0.7678$ \\
\midrule
-- & True model ($B_{\rm true}$) & -- & 0.0856 & $0.0938\pm0.0058$ & 0.1952 & $-0.8007$ \\
\bottomrule
\end{tabular}
\end{table}

Learning a useful shared prior requires enough independent training functions (Table~\ref{tab:proposed-multi-function-results}). For $M\leq16$, both the knee-selected and unregularised priors predict less accurately than the fixed source; at $M=1$, the knee-selected fit has latent RMSE $0.29$, compared with $0.11$ for the source. At $M=64$, their RMSEs are comparable, with the knee-selected fit improving on the source in 12 of 20 seeds. At $M=256$, the proposed knee-selected fit improves over the source in every seed on all three predictive metrics (paired Wilcoxon $p<10^{-5}$), and its mean latent RMSE is within $3\%$ of the true model's value ($0.0965$ versus $0.0938$).

\begin{figure}[!t]
\centering
\includegraphics[width=\textwidth]{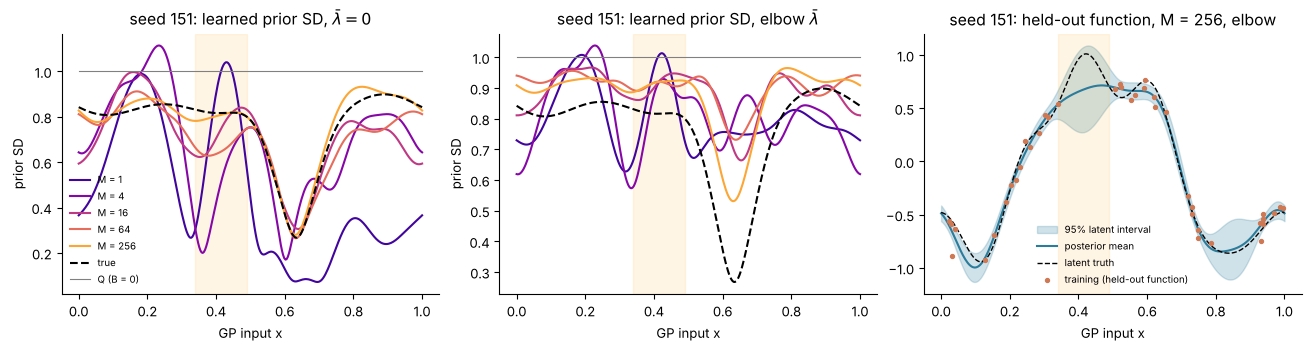}
\caption{Prior recovery as $M$ grows. Left/centre: prior standard deviations (SD) of unregularised/knee-selected fits; dashed: truth; grey: source. Right: held-out prediction at $M=256$ using the knee-selected prior, with posterior mean, latent truth, observations, and pointwise $95\%$ latent interval conditional on fitted parameters. Shading marks the common observation gap; intervals exclude noise.}
\label{fig:proposed-multi-prior}
\end{figure}

Kinetic regularisation improves all three predictive metrics relative to the unregularised fit at every $M$. The benefit is considerable with few training functions and modest at $M=256$, where latent RMSE decreases from $0.0971$ to $0.0965$. These predictive gains do not imply equally accurate recovery of the prior covariance: the selected penalty can suppress genuine spatial variation, as shown in the prior standard-deviation profiles in Figure~\ref{fig:proposed-multi-prior}.

Figure~\ref{fig:proposed-multi-prior} illustrates the distinction between prior recovery and conditional prediction for seed 151. The true prior has standard deviation varying from $0.27$ to $0.90$, with a sharp dip near $x=0.63$, whereas the source has constant standard deviation $1$. Without regularisation, the maximum pointwise error in prior standard deviation decreases from $0.66$ at $M=1$ to $0.07$ at $M=256$, and the dip is recovered for $M\geq16$. The knee-selected fits retain more of the source variance: at $M=256$, the dip reaches only $0.53$, with a maximum pointwise error of $0.27$. This is the over-smoothing discussed in Section~\ref{sec:proposed-path-selection}. Nevertheless, conditioning on observations gives an accurate prediction in the illustrated held-out example outside the gap. Within the gap, the interval widens and contains the latent function, although the posterior mean misses its peak. Additional seed realisations are shown in Appendix~\ref{sec:additional-seeds}.

%% file: sections/conclusion.tex

\section{Conclusion}
\label{sec:conclusion}
We introduced a semigroup-based construction of GP priors in which a learned linear generator transforms a fixed Gaussian source. The construction preserves Gaussianity and admits an explicit covariance evolution, retaining standard GP inference while allowing regularisation of the generating path through its kinetic energy.

Our experiments show that learning a shared prior requires sufficiently many independent function draws. Kinetic regularisation improves held-out prediction relative to unregularised learning and substantially reduces path energy; with 256 training functions, the selected prior outperforms the fixed source in every seed and achieves latent RMSE within $3\%$ of the true model. These predictive gains coexist with a limitation: selecting the knee of the fit--energy curve can suppress genuine nonstationary structure, so accurate prediction does not imply equally accurate covariance recovery.

Future work should develop weight-selection criteria that adapt to the number of independent training functions and evaluate the construction on real nonstationary data. The central distinction is between learning the endpoint prior and choosing how to reach it: observations constrain the former, while kinetic regularisation provides a criterion for the latter without requiring identification of the true generator.

%% file: sections/appendix.tex

\appendix
\renewcommand{\theHfigure}{appendix.\arabic{figure}}

\subsection*{Generative AI Use}

OpenAI's Codex and Claude were used to assist with language editing, \LaTeX~formatting, Python coding, and verification for consistency across the manuscript. The authors remain responsible for the accuracy and integrity of all submitted content.

\section{Computational details}
\subsection{Regularisation grid and knee selection}
\label{sec:weight-selection-details}
We fit a regularisation path over
$\bar\lambda_{\kin}\in\{0\}\cup\{10^{-3},10^{-2.75},\dots,10^{2}\}$.
At each weight, we retain the best fit obtained from two random
initialisations and warm starts from neighbouring weights.
To select the knee, we keep the Pareto-optimal fits in kinetic energy and training NLL per observation, rescale both quantities separately to $[0,1]$ using their minima and maxima over these fits, and choose the weight whose rescaled fit lies farthest below the chord joining the two extreme points of this front, measured by perpendicular distance. Selection uses training data only.

\subsection{Monte Carlo estimation and learning algorithm}
\label{sec:general-computation}

When the kinetic expectation in Section~\ref{sec:model-computation} is not available in closed form, we use the following Monte Carlo estimator.
Draw $Z_0^{(s)}\overset{\mathrm{iid}}{\sim}\N(m_0,C_0)$ and use
\begin{equation}
\widehat{\mathcal L}_{\kin}(\theta)
=\frac{1}{2S}\sum_{s=1}^S
\int_0^1\|A_\theta T_t^\theta Z_0^{(s)}\|_\cH^2\,\dd t
\approx\frac{1}{2S}\sum_{s=1}^S\sum_{q=1}^Q
w_q\|A_\theta T_{t_q}^\theta Z_0^{(s)}\|_\cH^2.
\label{eq:model-mc}
\end{equation}
The source draws, quadrature nodes $t_q$, and weights $w_q$ are held fixed during optimisation; each draw is used along its entire trajectory.
Here $S$ controls the Monte Carlo approximation, independently of the
number of observed functions. Algorithm~\ref{alg:semigroup-learning}
summarises fitting for a fixed regularisation weight.

\begin{algorithm}[htbp]
\caption{Learning a semigroup GP prior}
\label{alg:semigroup-learning}
\begin{algorithmic}[1]
\Require Data $\cD$; source law $\N(m_0,C_0)$; generator family
$A_\theta$; weight $\lambda_{\kin}$; noise floor $\sigma_{\min}>0$.
\Ensure Fitted parameters $\widehat\Theta$ and prior
$\GP(m_{\widehat\theta},k_{\widehat\theta})$.
\State Initialise admissible $\Theta=(\theta,\sigma_n)$ and a time quadrature rule.
\If{the kinetic expectation is not available in closed form}
\State Draw and fix $Z_0^{(1)},\ldots,Z_0^{(S)}\sim\N(m_0,C_0)$.
\EndIf
\Repeat
\State Compute $m_\theta$ and $k_\theta$ using~\eqref{eq:model-kernel}, and $K_\Theta$ using~\eqref{eq:model-marginal}.
\State Evaluate $\mathcal L_{\mathrm{NLL}}$ in~\eqref{eq:nll} by Cholesky factorisation.
\State Evaluate $\mathcal L_{\kin}$ in~\eqref{eq:model-objective} analytically or by~\eqref{eq:model-mc}, using time quadrature as needed.
\State Differentiate the loss in~\eqref{eq:model-objective} and update $\Theta$, respecting its constraints.
\Until{the stopping criterion is met}
\State \Return the best fitted $\widehat\Theta$ and its endpoint GP.
\end{algorithmic}
\end{algorithm}
\FloatBarrier

\subsection{Fourier implementation and optimisation}
\label{sec:implementation-details}

For numerical stability in the Fourier model of Section~\ref{sec:application-model}, we choose $R_\theta$ lower triangular and impose the following bounds on the rotation and damping matrices respectively:
\[
\|S_\theta\|_F<6,
\qquad
\|R_\theta\|_F^2<24.
\]
The kinetic energy is computed exactly via the Lyapunov equation of Section~\ref{sec:application-model}; as a check, it agrees with 20-point Gauss--Legendre quadrature to machine precision. The generator parameters and observation-noise standard deviation $\sigma_n>0.02$ are fitted using
objective~\eqref{eq:model-objective}, with 260
Adam~\citep{kingma2015} iterations followed by
up to 40 L-BFGS~\citep{liu1989} iterations.

\subsection{Synthetic data, fitting, and evaluation}
\label{sec:synthetic-specification}

The nonstationary target prior in Section~\ref{sec:proposed-experimental-setup} is obtained by transforming the fixed
Fourier source $c_0\sim\N(0,Q)$ from Section~\ref{sec:application-model}:
\[
f_{\mathrm{true}}(x)=\phi(x)^\top e^{B_{\mathrm{true}}}c_0,
\qquad
B_{\mathrm{true}}
=-\operatorname{diag}(0.08+0.025\nu_j^2)
-1.2vv^\top+0.45(uw^\top-wu^\top).
\]
Here $u$ and $w$ select the
$\sqrt2\cos(2\pi x)$ and $\sqrt2\sin(4\pi x)$ modes, respectively, and $v$
is the normalised Fourier projection of
$\exp\{5[\cos(2\pi(x-0.63))-1]\}$ on 256 equally spaced points.

For the training sets described in Section~\ref{sec:proposed-experimental-setup}, we use nested sets of independent coefficient draws $c_0^{(r)}\sim\N(0,Q)$. Observation noise is Gaussian with standard deviation $0.08$. Writing $\mathcal L_{\mathrm{NLL}}^{(r)}$ for the NLL of function $r$, the shared-prior objective is
\[
\frac1N\sum_{r=1}^M\mathcal L^{(r)}_{\mathrm{NLL}}
+\bar\lambda_{\kin}\mathcal L_{\kin},
\qquad N=40M,
\]
using the normalisation in Remark~\ref{rem:lambda-normalization}.

Evaluation uses 50 independent target functions per seed, each with 40 observations outside the training gap, noise standard deviation $0.08$, and 120 test inputs over $[0,1)$. Latent RMSE is computed over all test inputs; gap RMSE uses only inputs in $[0.34,0.49]$. Noisy NLPD is the mean marginal negative log predictive density of noisy test observations. Aggregation and variability summaries follow the conventions in Table~\ref{tab:proposed-multi-function-results}.

\clearpage

\section{Prior recovery across additional seeds}
\label{sec:additional-seeds}

\begin{figure}[!ht]
\centering
\includegraphics[width=\textwidth,height=0.65\textheight,keepaspectratio]{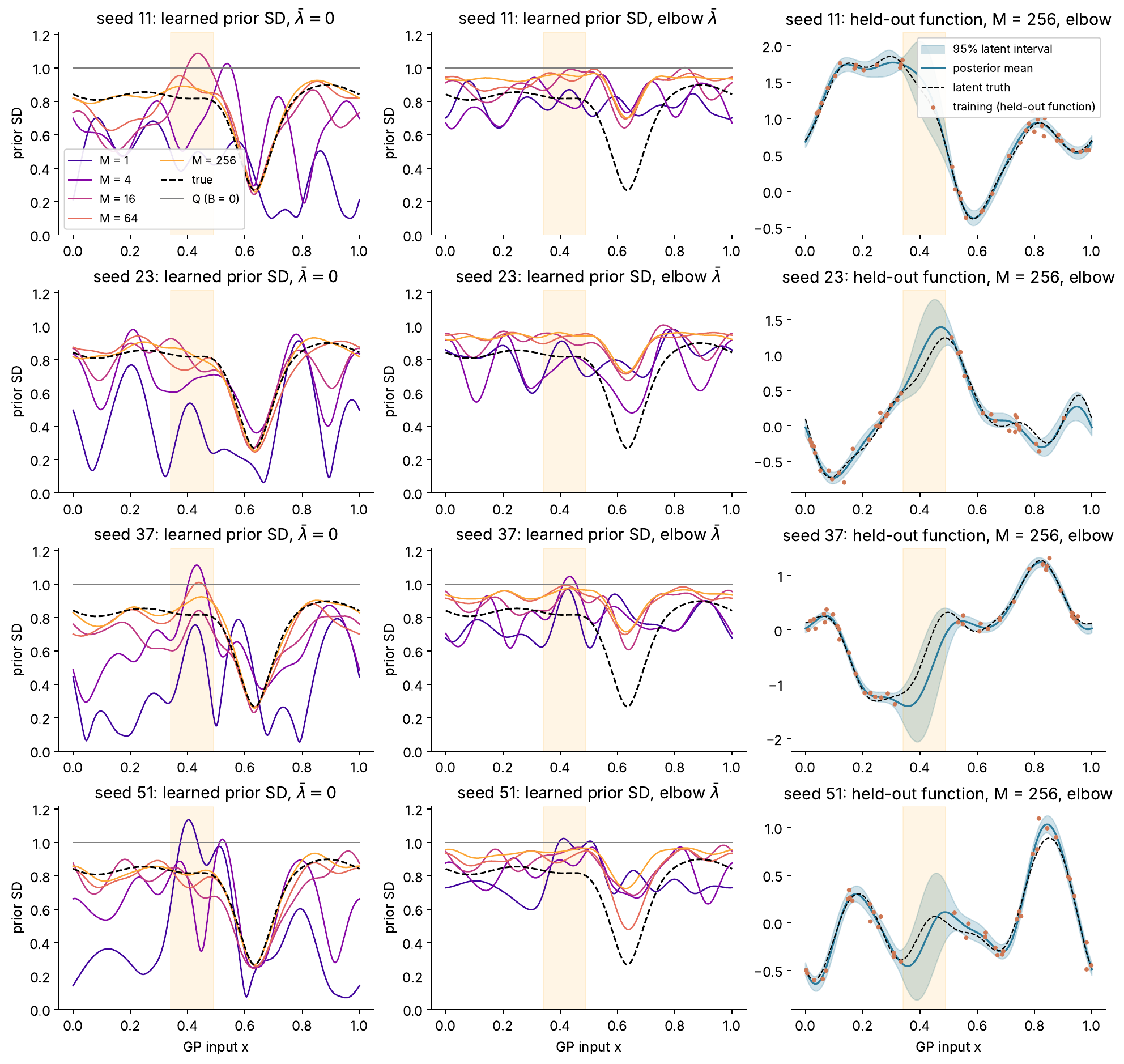}
\caption{Additional seeds: prior recovery and conditional prediction (group 1); conventions follow Figure~\ref{fig:proposed-multi-prior}.}
\label{fig:additional-seeds-1}
\end{figure}

\begin{figure}[p]
\centering
\includegraphics[width=\textwidth,height=0.85\textheight,keepaspectratio]{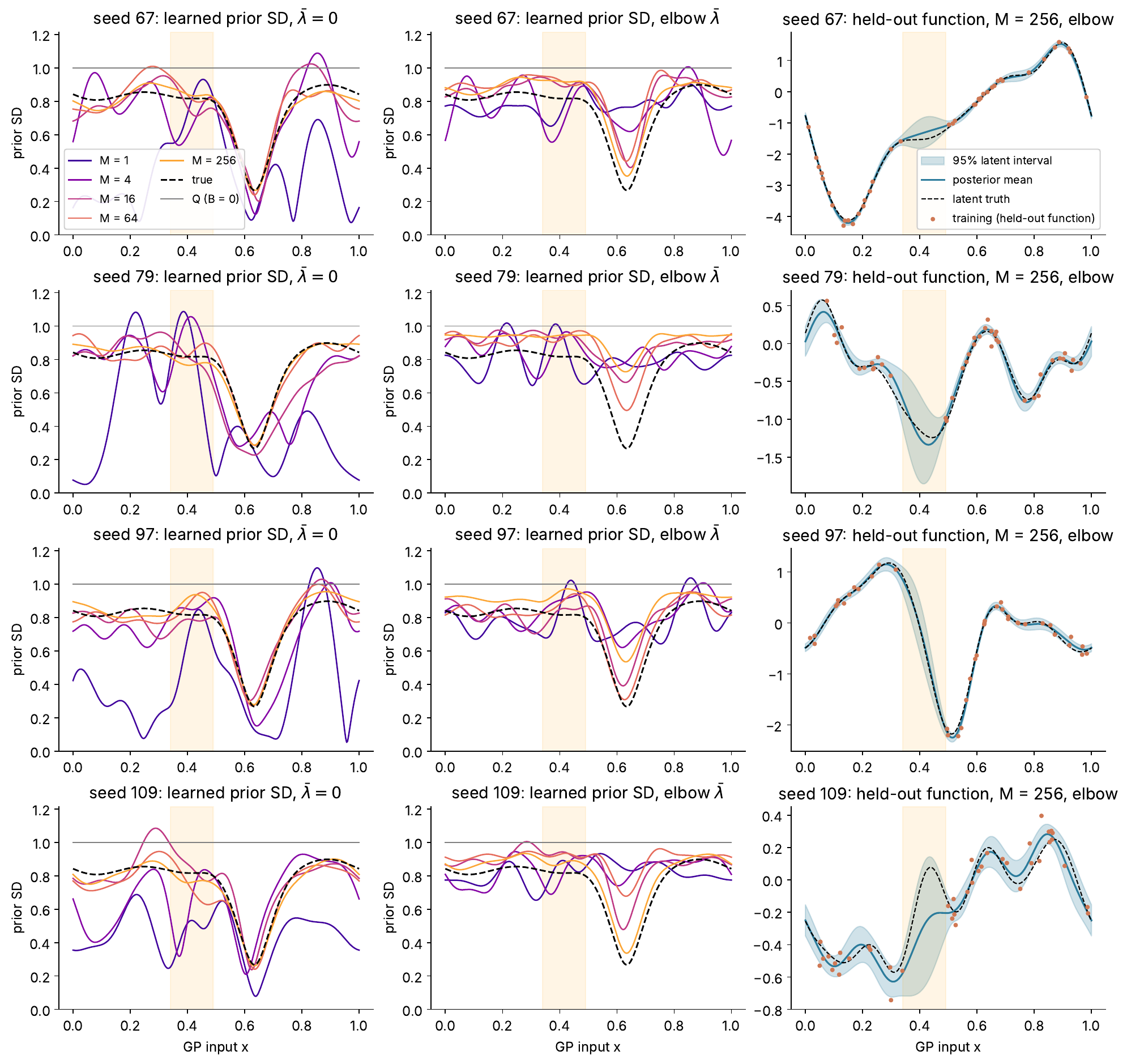}
\caption{Additional seeds: prior recovery and conditional prediction (group 2); conventions follow Figure~\ref{fig:proposed-multi-prior}.}
\label{fig:additional-seeds-2}
\end{figure}

\begin{figure}[p]
\centering
\includegraphics[width=\textwidth,height=0.85\textheight,keepaspectratio]{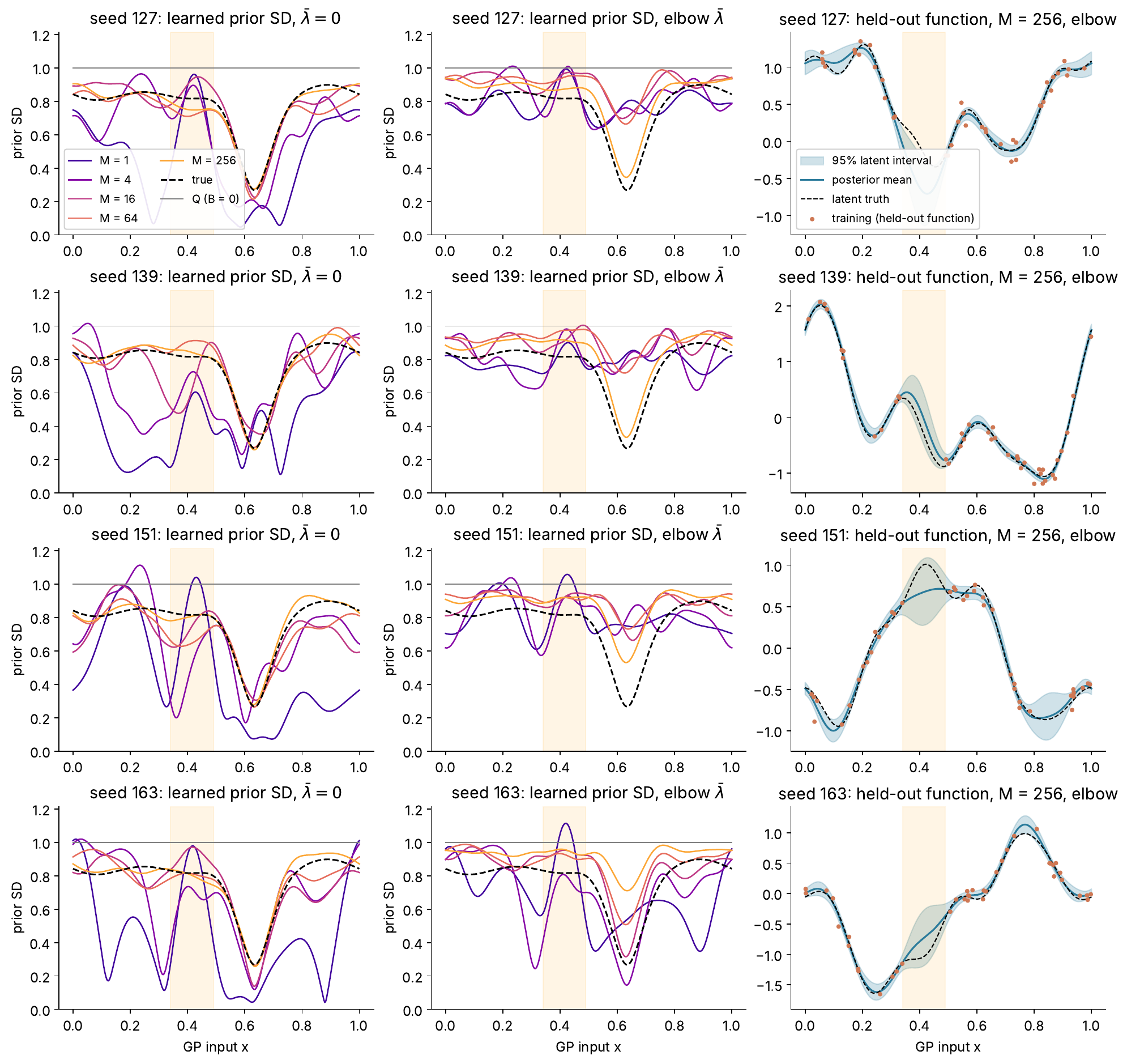}
\caption{Additional seeds: prior recovery and conditional prediction (group 3); conventions follow Figure~\ref{fig:proposed-multi-prior}.}
\label{fig:additional-seeds-3}
\end{figure}

\begin{figure}[p]
\centering
\includegraphics[width=\textwidth,height=0.85\textheight,keepaspectratio]{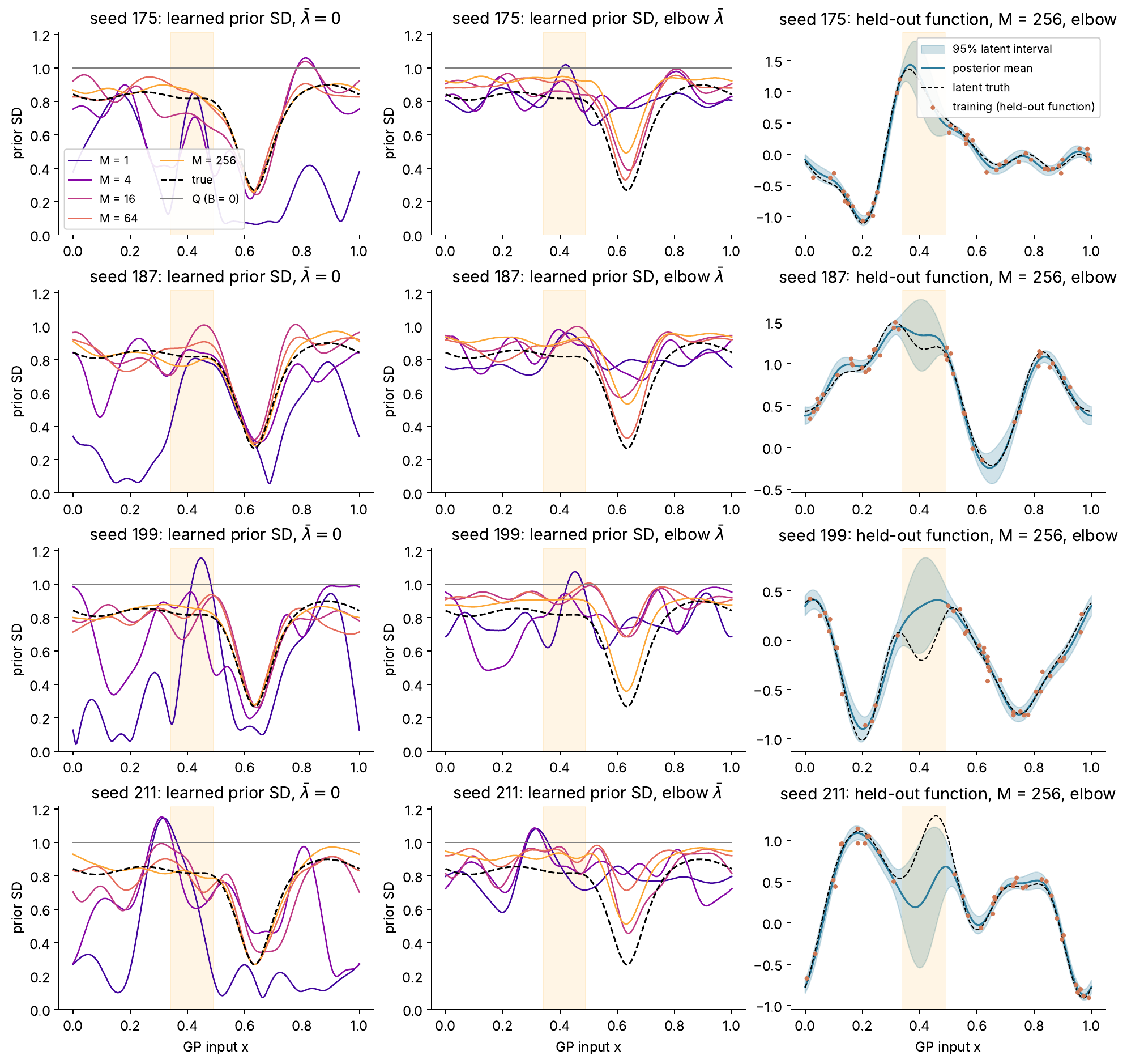}
\caption{Additional seeds: prior recovery and conditional prediction (group 4); conventions follow Figure~\ref{fig:proposed-multi-prior}.}
\label{fig:additional-seeds-4}
\end{figure}

\begin{figure}[!ht]
\centering
\includegraphics[width=\textwidth,height=0.85\textheight,keepaspectratio]{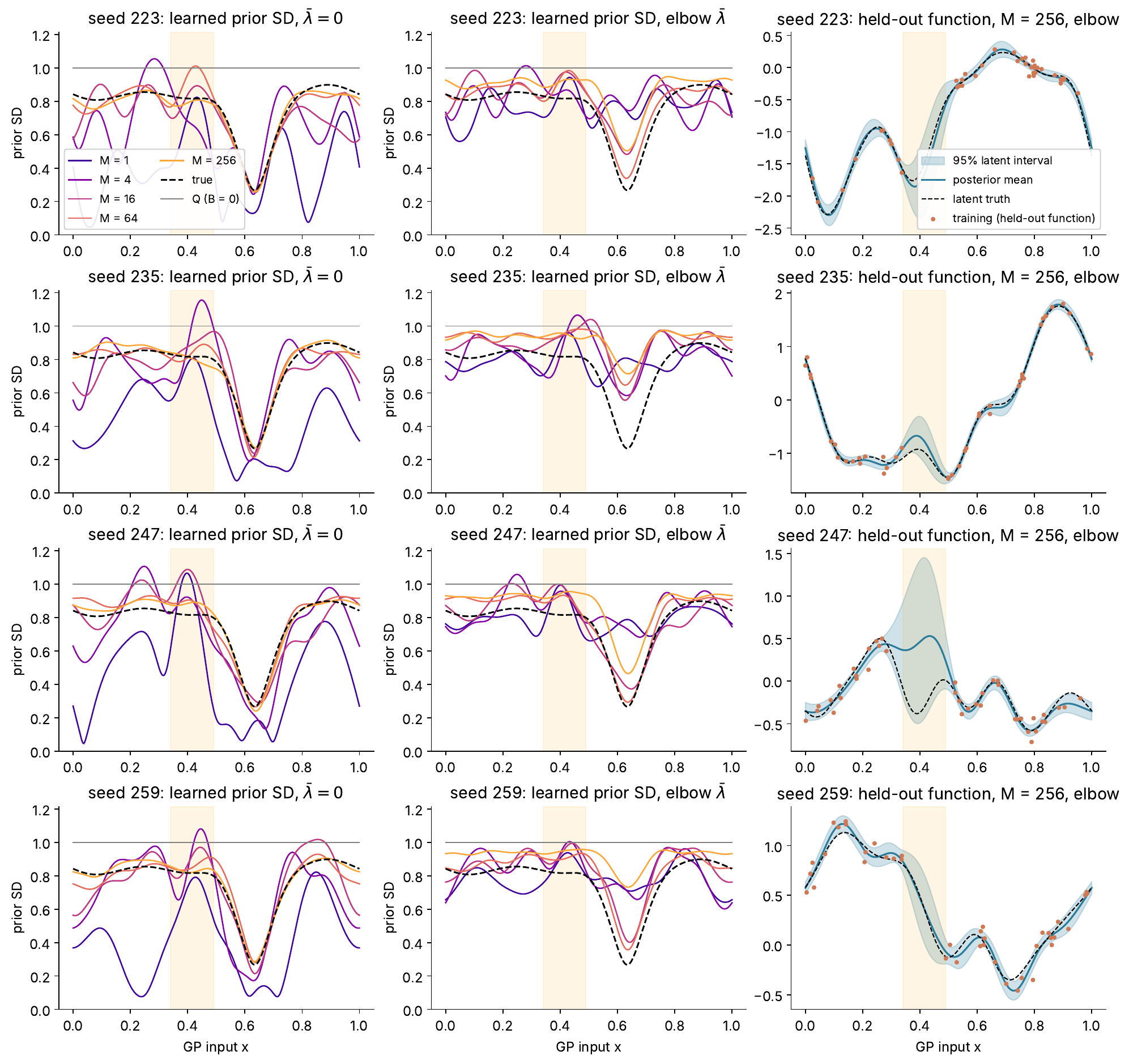}
\caption{Additional seeds: prior recovery and conditional prediction (group 5); conventions follow Figure~\ref{fig:proposed-multi-prior}.}
\label{fig:additional-seeds-5}
\end{figure}

Figures~\ref{fig:additional-seeds-1}--\ref{fig:additional-seeds-5} supplement Figure~\ref{fig:proposed-multi-prior} with prior-recovery and conditional-prediction plots across additional random seeds, using the experimental setup of Section~\ref{sec:proposed-experimental-setup}. 

\clearpage

\clearpage

%% file: main_arXiv.bbl
\begin{thebibliography}{19}
\providecommand{\natexlab}[1]{#1}
\providecommand{\url}[1]{\texttt{#1}}
\expandafter\ifx\csname urlstyle\endcsname\relax
  \providecommand{\doi}[1]{doi: #1}\else
  \providecommand{\doi}{doi: \begingroup \urlstyle{rm}\Url}\fi

\bibitem[Aronszajn(1950)]{aronszajn1950}
N.~Aronszajn.
\newblock Theory of reproducing kernels.
\newblock \emph{Transactions of the American Mathematical Society}, 68\penalty0
  (3):\penalty0 337--404, 1950.
\newblock \doi{10.1090/S0002-9947-1950-0051437-7}.

\bibitem[{Da Prato} and Zabczyk(2002)]{daprato2002}
G.~{Da Prato} and J.~Zabczyk.
\newblock \emph{{Second Order Partial Differential Equations in Hilbert
  Spaces}}.
\newblock Cambridge University Press, 2002.

\bibitem[Duvenaud et~al.(2013)Duvenaud, Lloyd, Grosse, Tenenbaum, and
  Ghahramani]{duvenaud2013}
D.~Duvenaud, J.~R. Lloyd, R.~Grosse, J.~B. Tenenbaum, and Z.~Ghahramani.
\newblock {Structure Discovery in Nonparametric Regression through
  Compositional Kernel Search}.
\newblock In \emph{Proceedings of the 30th International Conference on Machine
  Learning}, volume~28 of \emph{Proceedings of Machine Learning Research},
  pages 1166--1174, 2013.

\bibitem[Engel and Nagel(2000)]{engel2000}
K.-J. Engel and R.~Nagel.
\newblock \emph{{One-Parameter Semigroups for Linear Evolution Equations}}.
\newblock Springer, 2000.

\bibitem[Finlay et~al.(2020)Finlay, Jacobsen, Nurbekyan, and
  Oberman]{finlay2020}
C.~Finlay, J.-H. Jacobsen, L.~Nurbekyan, and A.~Oberman.
\newblock {How to train your neural ODE: the world of Jacobian and kinetic
  regularization}.
\newblock In \emph{Proceedings of the 37th International Conference on Machine
  Learning}, volume 119 of \emph{Proceedings of Machine Learning Research},
  pages 3154--3164, 2020.

\bibitem[Higdon(1998)]{higdon1998}
D.~Higdon.
\newblock {A process-convolution approach to modelling temperatures in the
  North Atlantic Ocean}.
\newblock \emph{Environmental and Ecological Statistics}, 5:\penalty0 173--190,
  1998.

\bibitem[Kingma and Ba(2015)]{kingma2015}
D.~P. Kingma and J.~Ba.
\newblock {Adam: A method for stochastic optimization}.
\newblock In \emph{Proceedings of the 3rd International Conference on Learning
  Representations}, 2015.

\bibitem[Kondor and Lafferty(2002)]{kondor2002}
R.~I. Kondor and J.~Lafferty.
\newblock Diffusion kernels on graphs and other discrete input spaces.
\newblock In \emph{Proceedings of the 19th International Conference on Machine
  Learning}, pages 315--322, 2002.

\bibitem[Lindgren et~al.(2011)Lindgren, Rue, and Lindstr{\"o}m]{lindgren2011}
F.~Lindgren, H.~Rue, and J.~Lindstr{\"o}m.
\newblock {An explicit link between Gaussian fields and Gaussian Markov random
  fields: the stochastic partial differential equation approach}.
\newblock \emph{Journal of the Royal Statistical Society: Series B},
  73\penalty0 (4):\penalty0 423--498, 2011.

\bibitem[Liu and Nocedal(1989)]{liu1989}
D.~C. Liu and J.~Nocedal.
\newblock {On the limited memory BFGS method for large scale optimization}.
\newblock \emph{Mathematical Programming}, 45:\penalty0 503--528, 1989.

\bibitem[Paciorek and Schervish(2006)]{paciorek2006}
C.~J. Paciorek and M.~J. Schervish.
\newblock Spatial modelling using a new class of nonstationary covariance
  functions.
\newblock \emph{Environmetrics}, 17\penalty0 (5):\penalty0 483--506, 2006.

\bibitem[Rasmussen and Williams(2006)]{rasmussen2006}
C.~E. Rasmussen and C.~K.~I. Williams.
\newblock \emph{{Gaussian Processes for Machine Learning}}.
\newblock MIT Press, 2006.

\bibitem[Remes et~al.(2017)Remes, Heinonen, and Kaski]{remes2017}
S.~Remes, M.~Heinonen, and S.~Kaski.
\newblock Non-stationary spectral kernels.
\newblock In \emph{Advances in Neural Information Processing Systems},
  volume~30, pages 4645--4654, 2017.
\newblock URL \url{https://arxiv.org/abs/1705.08736}.

\bibitem[Riedle(2014)]{riedle2014}
M.~Riedle.
\newblock Stochastic integration with respect to cylindrical {L\'evy} processes
  in {Hilbert} spaces: an {$L^2$} approach.
\newblock \emph{Infinite Dimensional Analysis, Quantum Probability and Related
  Topics}, 17\penalty0 (1):\penalty0 1450008, 2014.
\newblock \doi{10.1142/S0219025714500088}.

\bibitem[Saat{\c{c}}i et~al.(2010)Saat{\c{c}}i, Turner, and
  Rasmussen]{saatci2010}
Y.~Saat{\c{c}}i, R.~D. Turner, and C.~E. Rasmussen.
\newblock {Gaussian Process Change Point Models}.
\newblock In \emph{Proceedings of the 27th International Conference on Machine
  Learning}, pages 927--934, 2010.

\bibitem[Solin and S{\"a}rkk{\"a}(2020)]{solinsarkka2020}
A.~Solin and S.~S{\"a}rkk{\"a}.
\newblock {Hilbert space methods for reduced-rank Gaussian process regression}.
\newblock \emph{Statistics and Computing}, 30:\penalty0 419--446, 2020.

\bibitem[Tobar et~al.(2015)Tobar, Bui, and Turner]{tobar2015}
F.~Tobar, T.~D. Bui, and R.~E. Turner.
\newblock {Learning stationary time series using Gaussian processes with
  nonparametric kernels}.
\newblock In \emph{Advances in Neural Information Processing Systems},
  volume~28, 2015.

\bibitem[Wilson and Adams(2013)]{wilson2013}
A.~G. Wilson and R.~P. Adams.
\newblock {Gaussian Process Kernels for Pattern Discovery and Extrapolation}.
\newblock In \emph{Proceedings of the 30th International Conference on Machine
  Learning}, volume~28 of \emph{Proceedings of Machine Learning Research},
  pages 1067--1075, 2013.

\bibitem[Wilson et~al.(2016)Wilson, Hu, Salakhutdinov, and Xing]{wilson2016}
A.~G. Wilson, Z.~Hu, R.~Salakhutdinov, and E.~P. Xing.
\newblock {Deep Kernel Learning}.
\newblock In \emph{Proceedings of the 19th International Conference on
  Artificial Intelligence and Statistics}, volume~51 of \emph{Proceedings of
  Machine Learning Research}, pages 370--378, 2016.

\end{thebibliography}
